\documentclass[runningheads]{llncs}

\usepackage{graphicx}
\usepackage{hyperref}
\usepackage{float} 
\usepackage{tikz}
\usepackage{makecell}
\usepackage{tabu}
\usepackage{textcomp}
\usepackage{multirow}
\usepackage{paralist}
\usepackage{booktabs}
\usepackage{adjustbox}
\usepackage{url}

\begin{document}

\title{Credibility-based Fake News Detection}

\author{Niraj Sitaula \and
Chilukuri K. Mohan \and
Jennifer Grygiel \and \\
Xinyi Zhou \and
Reza Zafarani}
\authorrunning{N. Sitaula et al.}
%
\institute{ 
Syracuse University, Syracuse, NY 13244, USA \\
\email{\{nsitaula,mohan,jgrygiel\}@syr.edu}\\
\email{\{zhouxinyi,reza\}@data.syr.edu}}
\maketitle              

\begin{abstract}
Fake news can significantly misinform people who often rely on online sources and social media for their information.  
Current research on fake news detection has mostly focused on analyzing fake news content and how it propagates on a network of users. 
In this paper, we emphasize the detection of fake news by assessing its credibility. 
By analyzing public fake news data, we show that information on news sources (and authors) 
 can be a strong indicator of credibility. 
 Our findings suggest that an author's history of association with fake news, and the number of authors of a news article, can play a significant role in detecting fake news.
Our approach can help improve traditional fake news detection methods, wherein 
content features 
 are often used to detect fake news.
 
\end{abstract}

\section{Introduction}
In this digital age, news and information are mostly received from various online sources. Surveys have shown that a large number of online users depend on social media for the news: 51\% in Australia, 46\% in Italy, 40\% in the United States, and 36\% in the United Kingdom~\cite{fletcher2018people}.  Hence, fake news can misinform many people who rely on online news and/or social media for the information.

The impact of fake news has been widely discussed after the 2016 U.S. presidential election.  A study by Silverman~\cite{silverman2016analysis} shows that for the top 20 election stories in 2016, the top 20 fake news stories had 8,711,000 shares, reactions, and comments on Facebook. These user engagement numbers were significantly higher than those for 
the top 20 real stories, with 7,367,000 shares, reactions, and comments on Facebook during the same time period. 
These concerns motivate assessing news credibility and detecting fake news \textit{before} it spreads online.

Detecting fake news has gained attention from many academic researchers as well as other organizations. 
Although some fact checking websites exist, such as FactCheck\footnote{\url{https://www.factcheck.org}} and PolitiFact\footnote{\url{https://www.politifact.com/}}, the problem of detecting fake news 
is far from being 
solved. 
Manually verifying each and every fact in the news is extremely difficult with the high volume of data being created and shared every minute. Furthermore, it has become extremely difficult 
to decide whether a news article is fake or credible, since 
fake news articles often contain false information as well as some facts. 
Potthast, et al.~\cite{potthast2017stylometric}, observe 
that fake news articles may contain facts,
and  credible news articles may contain factual errors.
Hence, an automated process to detect fake news based only on  content verification may not be 
effective. 
If we 
emphasize 
past information about the sources (authors or URLs) of news articles, then deception 
is still possible with new URLs and new fake author names. 
Our goal 
is to identify general indicators of news credibility,
using both (1) \textit{source} and (2) \textit{content} perspectives to help detect fake news. 

Using public data for fake news detection~\cite{shu2018fakenewsnet}, we have analyzed multiple features of
information related to the sources and contents of news articles. Our analyses demonstrate that fake news can be distinguished from true news based on features related to source and content. We also observe that while some features exhibit differences between fake and factual news, they do not help better predict fake news. 

This paper focuses on finding signals or indicators of news credibility that can help detect fake news. Our findings suggest that the information about authors of news articles
can indicate news credibility and help 
detect fake news. 
Using only information on the number of authors 
and the authors' publication history, classifiers were able to obtain $>$0.75 average $F_1$-score. When content related features were added to these features, we observed further improvements when detecting fake news. 


In the following, we detail our analysis on various aspects of credibility. We review related work in Section \ref{sec:related}, followed by a brief description of the dataset used for analysis in Section \ref{sec:datasets}. Section \ref{sec:source} provides our analysis of credibility based on the source of the news and Section \ref{sec:content} details credibility factors based on the content of news. Based on our analysis on source and content credibility, we build predictive models to detect fake news, which we detail in Section \ref{sec:result}. We conclude in Section \ref{sec:conclusion} and present some directions for future work.

\section{Related Work}
\label{sec:related}
We briefly discuss research on fake news detection and credibility assessment.\vspace{1mm}



\noindent \textbf{Fake News Research.} Fake news has been an active area of research, where is has been detected often by relying on (1) news content and/or (2) social context information. Often formed from text and images along with news sources (authors or websites), news content has been utilized in various ways to detect fake news. 
Text has been represented as a set of \texttt{(subject, predicate, object)} 
features and used to predict fake news by developing link prediction algorithms, i.e., how likely the extracted \texttt{predicate} connects the \texttt{subject} with the specific \texttt{object}~\cite{hassan2017toward,ciampaglia2015computational,shi2016discriminative}. Such textual information can be represented as style features at various language levels as well, e.g., lexicon-level~\cite{zhou2019fake,perez2017automatic,wang2018eann,zhang2018fake}, syntax-level~\cite{feng2012syntactic,zhou2019fake}, semantic-level~\cite{perez2017automatic}, and discourse-level~\cite{rubin2015truth,karimi2019learning}, based on  $n$-grams~\cite{perez2017automatic}, Term Frequency–Inverse Document Frequency (TF-IDF)~\cite{perez2017automatic}, Bag-Of-Words (BOWs)~\cite{zhou2019fake}, \textsc{word2vec}~\cite{zhou2019fake,mikolov2013efficient}, Part-Of-Speeches (POSs)~\cite{feng2012syntactic,zhou2019fake,horne2017just}, Context Free Grammers (CFGs)~\cite{feng2012syntactic,perez2017automatic,zhou2019fake}, Linguistic Inquiry and Word Count (LIWC)~\cite{perez2017automatic}, rhetorical relationships among sentences~\cite{rubin2015truth,karimi2019learning}, etc.;
features can be explicit, i.e., non-latent features such as the frequencies of lexicons, or implicit, i.e., latent features obtained by, for example, \textsc{word2vec}~\cite{wang2018eann,zhang2018fake}). 
Recently, news images and source websites have 
been 
used in fake news analyses. For example, to investigate news images, Jin et al.~\cite{jin2017novel} defined a set of visual features to predict fake news within a traditional statistical learning framework, and Wang et al.~\cite{wang2018eann} employed a deep neural network (VGG-19) to help extract the latent representation of news images. Baly et al.~\cite{biasOfNews} characterized fake news articles by their source websites, e.g., if they have a Wikipedia page, if their URLs contain digits or domain extensions such as {\tt .co}, {\tt.com}, {\tt.gov}, and their Web traffic information.   Nevertheless, few research 
efforts have focused on the authors who create and write the [true or fake] news, which we  investigate in this paper. 

On the other hand, fake news detection models have emerged in recent years by studying how fake news propagates on social media (i.e., using social context information). An example can be seen in the work by Vosoughi et al.~\cite{vosoughi2018spread}, which revealed that fake news spreads faster, farther, more widely, and is more popular compared to true news. Currently, methods to predict fake news have investigated the profiles of users
~\cite{scienceAdvance}, 
their social connections~\cite{zhou2019network,shu2019beyond}, and posts. For instance, Guess et al. ~\cite{scienceAdvance}
found that the age of a user is an important indicator of the frequency that he or she engages in fake news activities; the analysis showed that users over 65 shared fake news approximately seven times more often compared to the younger age group. While remarkable progress has been made to achieve early detection of fake news, 
very little social context information on news propagation 
may be available; this motivates the development of approaches that can detect fake news by focusing mainly on news content.

A comprehensive survey of the various approaches for handling fake news problem 
is given in the work by Zhou and Zafarani~\cite{fakeNewsSurvey}.
\vspace{1mm}


\noindent \textbf{Credibility Assessment.} The credibility of information (including news) is often evaluated by its quality and believability~\cite{castillo2011information}.
Research has specifically focused on assessing the credibility of social context information.
TweetCred, a real-time system, scores tweet credibility by using a set of hand-crafted features within a semi-supervised learning framework~\cite{gupta2014tweetcred}. Out of all selected features, the most important ones include the number of (unique) characters and words in tweets, indicating the significant correlation between the content of tweets and their credibility.
In another relevant study, Gupta, et al.~\cite{gupta2012twitter} found that a majority of the content generated at the time of crisis are from unknown sources (users) and at the same time, rumors are spread, which emphasized the importance of sources on information credibility. 
Based on a binary classifier, Castillo et al.~\cite{castillo2011information} evaluated information credibility on Twitter using hand-crafted features from users' posting and re-posting behavior, from the text of the posts, and from citations to external sources, which can achieve a precision and recall value between 0.7 and 0.8.
Within a similar framework, O'Donovan, et al.~\cite{odonovan2012credibility}, discovered 
that features 
such as URLs, mentions, retweets, and tweet length were among 
the best indicators of the credibility using eight diverse Twitter datasets.  Morris, et al.~\cite{morris2012tweeting}, found that the name of 
a Twitter user 
and the use of standard grammar, influence the credibility 
of tweets. 

In 
our work, we have adopted features from earlier findings mostly from the context of microblogging sites such as Twitter, to find how closely they relate to news content credibility.
We have not found 
any credibility study focused on number of authors related to the news, authors' collaboration relationships, and authors' past association with fake news articles.
The analysis and findings of this paper
provide insights to 
address these issues and improve fake news identification efforts.

\begin{figure}[t]
	\centering
	\includegraphics[width=\textwidth]{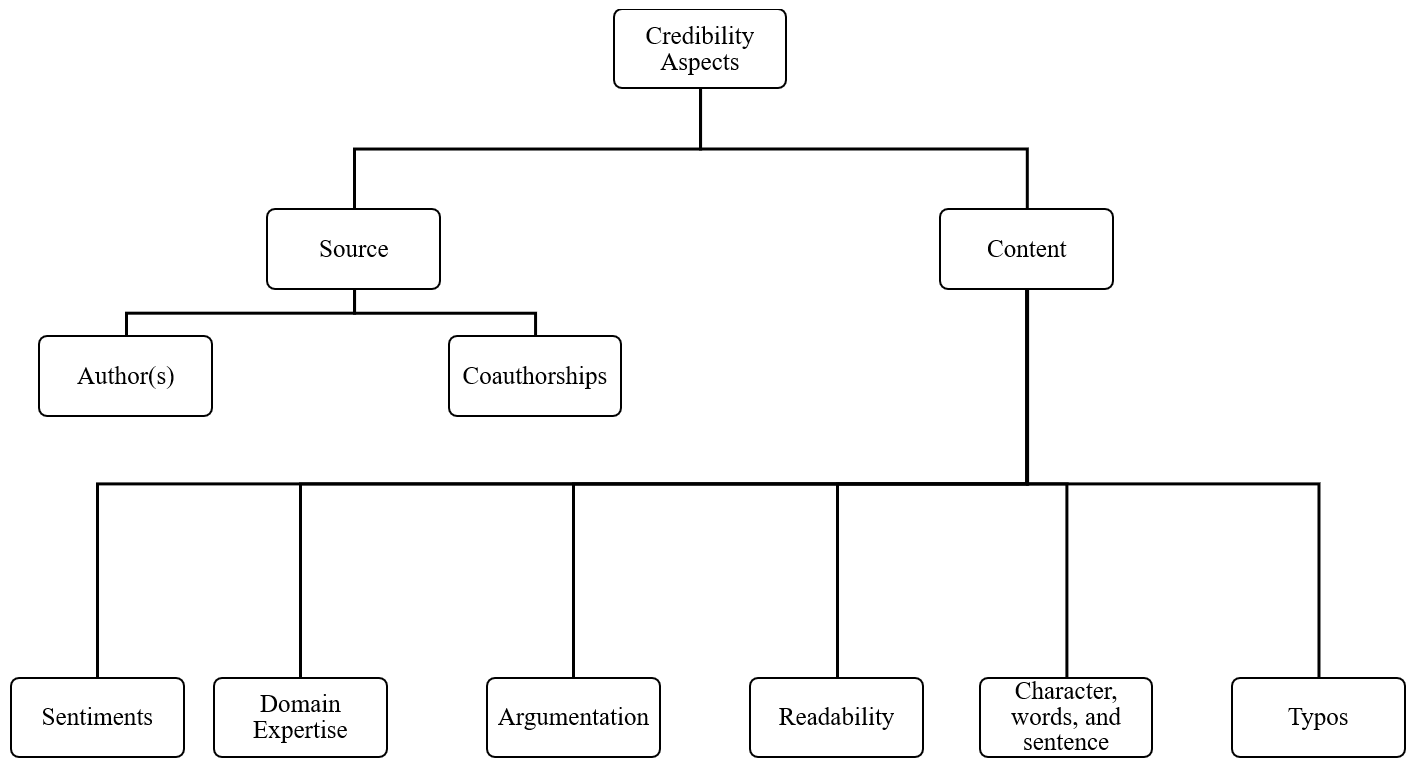}
	\caption{Hierarchical Structure of Credibility Aspects}
	\label{fig:hierarchy}
\end{figure}

We assess credibility from two broad perspectives: (i) Source and (ii) Content which are discussed in details in Sections \ref{sec:source} and \ref{sec:content}. For each category, we identify information in fake news that can capture 
various aspects of credibility, as shown in Figure \ref{fig:hierarchy}. Before further elaboration of these categories, we briefly 
describe the datasets used in our study. 

\section{Experimental Data}
\label{sec:datasets}

We have used two public datasets for fake news detection, 
 from {\url{https://bit.ly/2mHGnBI}}~\cite{shu2017fake,shu2019beyond,zhou2019fake,zhou2019network}. These datasets are from Buzzfeed news and Politifact. Besides news content and news labels (i.e., {\it fake} or {\it true}), the datasets contain 
information on
the social networks of users involved in spreading the news. Statistics of 
the two datasets are provided in Table~\ref{tab:datasets}.

\begin{table}[t]
\centering
\caption{Data Statistics}
\label{tab:datasets}
\begin{tabular}{|l|r|r|}\hline
\multicolumn{1}{|l|}{\textbf{Data}} & \multicolumn{1}{c|}{\textbf{PolitiFact}} & \multicolumn{1}{c|}{\textbf{BuzzFeed}} \\ \hline\hline
 \# Users & 23,865 & 15,257 \\ 
 \# News--Users & 32,791 & 22,779 \\ 
 \# Users--Users & 574,744 & 634,750 \\
 \# News Stories & 240 & 182 \\ 
 \# True News & 120 & 91 \\ 
 \# Fake News & 120 & 91 \\\hline

\end{tabular}
\end{table}

Out of 422 news articles in both datasets, 16 news articles had the same content/text, and 
were excluded from our analysis. The datasets were processed using {\it pandas}~\cite{mckinney2010data}, and {\it matplotlib}~\cite{hunter2007matplotlib} was used for visualization. In the following sections, we will discuss how different aspects of credibility can be captured from such data and demonstrate our findings on these datasets.

\section{Source Credibility}
\label{sec:source}
In this section, we present our analysis to derive credibility from the news source, i.e. news URL, number of authors of the news, coauthorship relation to credibility, author(s) affiliations, and history of credibility of authors.

Earlier research has focused on assessing source credibility by looking at the URL associated with any
news article~\cite{biasOfNews}, 
including features such as whether a website contains the \texttt{https} prefix, 
numbers,  
or {\tt .gov, .co, .com} domain extensions.
In our data, 354 news articles used the {\tt http} prefix, 15 used the {\tt https} prefix, and 37 had no URL. Out of the 354 URLs with {\tt http} prefix, only 154 were fake news. 
Surprisingly, 14 out of 15 news articles with the {\tt https} prefix were fake. 
These observations contradict past studies: 
having {\tt https} in a URL
does 
not 
imply credibility or help differentiate fake news from true news. 

Other studies~\cite{gupta2014tweetcred,gupta2012twitter,castillo2011information,morris2012tweeting} have shown that specific users information can be good indicators of credibility on Twitter. Hence, 
we seek generic user information that can capture credibility and help detect fake news. We group such {\it source} information into information on: 

\begin{compactenum}
\item \textit{Author(s)} of the news articles, and 
\item \textit{Coauthorships}, i.e., author collaborations. 
\end{compactenum}
We now discuss each of these subcategories in 
detail.

\subsection{Credibility Signals in Author(s)}

If a news article does not provide any information on its authors, 
its credibility can be questioned. 
An earlier study found that rumors mostly spread during the times of crises on Twitter and the majority of such rumors are posted by unknown sources/users~\cite{gupta2012twitter}. However, having the name(s) of the author(s)
is insufficient, because 
fake names or fake profiles can be easily created. Previous work has also looked at whether the news source is Wikipedia or a verified social media account, or contains other attributes to verify its credibility~\cite{biasOfNews}. 

\begin{figure}[t]
	\centering
	\includegraphics[width=0.8\textwidth]{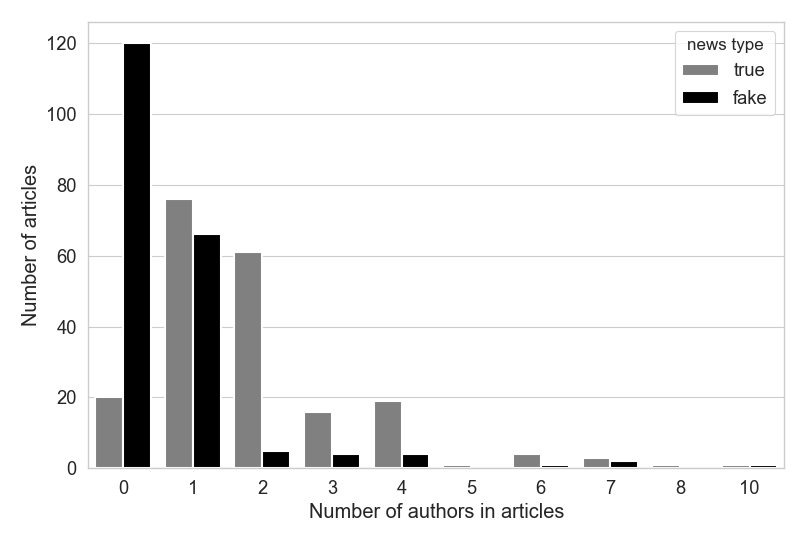}
	\caption{Frequency of number of authors in fake and true news. If an article has more than one author, it is more likely to be true news.}
	\label{fig:hist}
\end{figure}

Thus, credibility assessment methods require multiple steps. 
First, we simplify credibility assessment to focus on the number of authors in the two types of news, i.e., whether it contains no authors, one author, two authors, or more. 
We found that the average number of authors
is 0.66
for fake news  
and 1.97 for true news. The median number of authors is 0 for fake news and 2 for true news. Figure \ref{fig:hist} provides the distribution of the number of authors for true and fake news.

From Figure \ref{fig:hist}, we observe that the number of authors of a news article does have 
some correlation to its credibility. 
If an article has more than one author, it is more likely to be credible, and news with no author name is more likely to be fake news. The Pearson correlation coefficient between labels (true/fake) and number of authors is $0.406$. It is difficult to draw similar inferences when news articles have only one author. From the figure, we observe that there are almost equal proportions of fake and true news for articles with a single author.  A $p$-value of $<0.05$ was obtained after running Shapiro-Wilk normality test for the number of authors, which indicates that the distribution is not normal. As the distributions are not normal, we cannot compute the significance in differences in mean values between fake news and true news.  Hence, we used  Mann-Whitney U test on number of authors on two types of news, i.e., fake and true news. The $p$-value of $<0.001$ shows that the median number of authors in these two types of news can capture credibility. 
In our later analyses, we discuss ways to add past association of the authors with fake news to tackle the case when there is only one author. 

\subsection{Credibility Signals in Coauthorships}

Earlier studies on coauthorship networks (e.g., Newman~\cite{newman2004coauthorship}) found that (1) a small number of influential individuals exist in such networks and (2) disconnecting such individuals from the network can result in a set of small disjoint networks. Such observations motivate us to explore whether such influential authors exist in the network of news authors, and if they do, can they help assess the credibility of the news, and their coauthors. Hence, we extended our analysis by looking at the network of news authors and classifying them into three groups: authors who are (i) only associated with fake news, (ii) only associated with true news, and (iii) associated with both fake and true news. The objective is to analyze news credibility, given the \textit{position} of the author in this network as well as its \textit{neighbors} (other authors). 
This approach allows us to understand whether fake news authors collaborate only with other fake news authors, or if they also collaborate with true news authors. We raise similar questions for true news authors and those who publish both.  

Among the 237 unique authors in our data, 87 authors were authors of at least two or more news articles. To have sufficient historical data, authors 
whose names occurred only once (in our data) were excluded from the analysis. For simplicity, we only considered news articles whose authors were in the set of 87 authors. To provide clear insights on coauthorships, we assign these 87 authors to one of three groups:

\begin {enumerate}
\item {\it True-news authors}, only associated with two or more true news stories;
\item {\it Fake-news authors}, only associated with two or more fake news stories;
\item {\it Fake+True authors}, who have published both fake news and true news.
\end{enumerate}

\begin{figure}[t]
	\centering
	\includegraphics[width=0.75\textwidth]{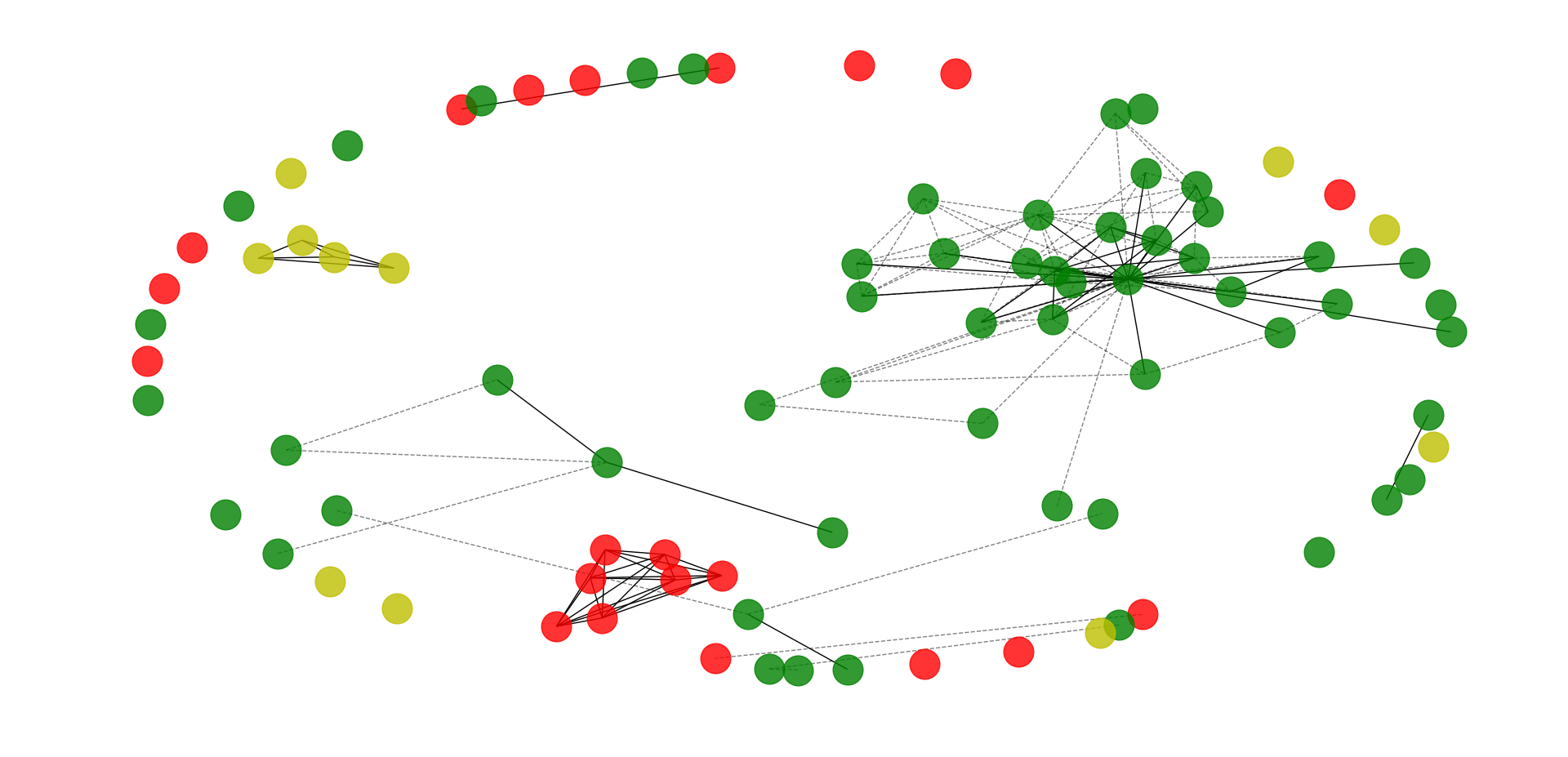}
	\caption{Authors Collaboration Network, showing
	authors that publish only fake news (red), only true news (green), or both (yellow).
	}
	\label{fig:authNetwork}
\end{figure}

For these groups, the coauthorship network among authors is shown
in Figure \ref{fig:authNetwork}, where green nodes represent {\it True-news authors}, red nodes represent {\it Fake-news authors}, and yellow nodes represent {\it Fake+True authors}. Dashed lines connect authors that have collaborated only once, whereas solid lines connect authors who have collaborated more than once. We notice that only 12.7\% of 87 authors are involved in both fake and true news, whereas the majority were either exclusively involved in fake news or true news. We also observe that fake news authors are 
often either the only author (of the fake news article), or are more likely to collaborate with other {\it Fake-news authors} (rather than with 
{\it True-news authors} or {\it Fake+True authors}). We had similar observations for {\it True-news authors}  (green nodes in Figure \ref{fig:authNetwork}) and  {\it Fake+True authors} (yellow nodes in Figure \ref{fig:authNetwork}). 

\begin{figure}[t]
	\centering
	\includegraphics[width=0.8\textwidth]{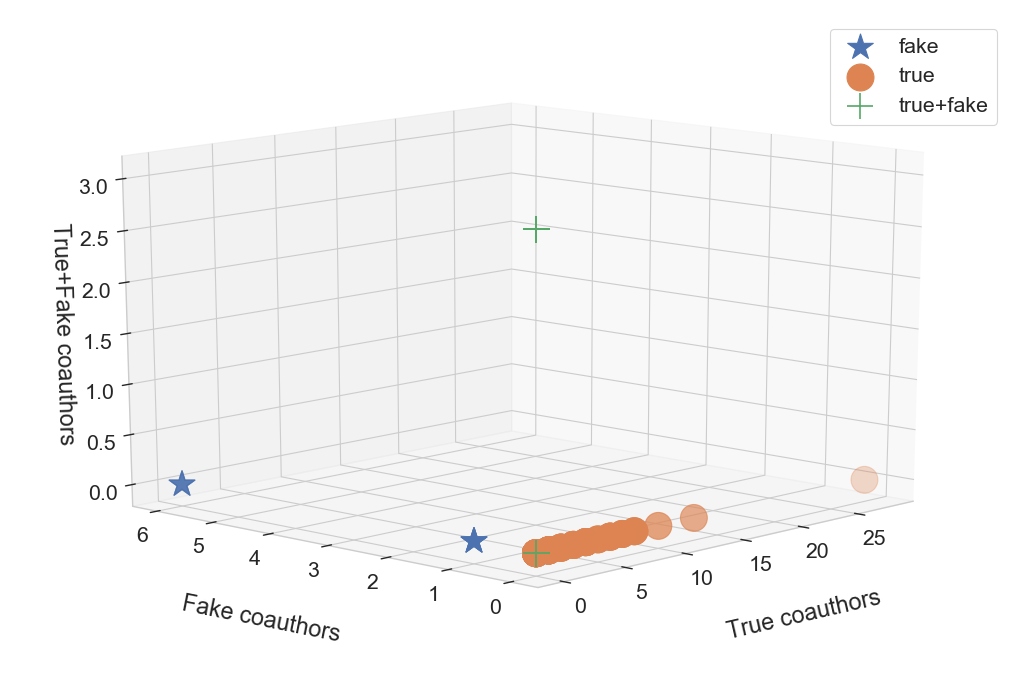}
	\caption{3D plot of authors. An author is represented in terms of three values: the number of coauthors of him or her who only post true news, fake news, or both.}
	\label{fig:3d}
\end{figure}

To further investigate these observations,
for each author in the coauthorship graph, we compute the number of coauthors (i.e., graph neighbors) 
who only post true news, only post fake news, and those who post both. Using these three numbers, we can represent any author as a 3D point and plot all authors in 3D space, as shown in Figure \ref{fig:3d}. We observe that the credibility of authors who have had multiple coauthorhips are easily distinguishable, as they often collaborate with the same type of authors. Hence, knowing the author's credibility, we can infer the credibility of  
coauthors. For authors with no neighbors (i.e., coauthors), they are indistinguishable. 

In sum, \textit{homophily}
exist in authorship~\cite{zafarani2014social}, where authors who write only true news are less likely to collaborate with authors who write fake news. These observations also indicate that if there are groups of authors associated with some news, by knowing credibility of any author, we may be able to infer the credibility of the news and its other authors. But, \textit{how can we determine an author's credibility?}  
Two observations 
help us address this question:\vspace{1mm}

\noindent \textbf{I. Affiliations provide information on credibility.} 
Some author names 
are associated with known organizations such as {\tt ABC news}, {\tt Associated Press}, {\tt Politico}, and {\tt CNN}. 
Hence, affiliations 
of authors 
with well-recognized organizations 
may indicate that a news article is not fake. 
While we did not explicitly consider the author relationships with the organizations, these findings support
the hypothesis that working for 
a credible organization allows one to infer author's credibility. Similarly, we found some unrealistic author names such as {\tt Fed up}, {\tt Ny evening}, {\tt About the Potatriot}, and {\tt About Stryker} associated with fake news. This finding corroborates the earlier observation by Gupta, et al.~\cite{gupta2012twitter},  where rumors 
on Twitter 
were shown to have been spread 
by unknown sources/users. \vspace{1mm}

\noindent \textbf{II. Historical record provides information on credibility.} The 87 authors that were selected were related to 172 news articles. In Table \ref{tab:distAuthors}, we looked at how different types of authors relate to the two types of news. We observe that around 28\% of the news articles have authors who post both fake and true news. However it is unclear how this information can help infer credibility of the authors or the news. To tackle this issue, we looked into 
the history of authors' credibility, i.e., their past associations with true/fake news articles, 
in order 
to explore whether 
these can capture the credibility 
of other articles authored later by the same authors.

\begin{table}[t]
\centering
\caption{Distribution of count of news articles based on coauthors.}
\begin{tabular}{|c|r|c|}\hline
\makecell{\bf News Type} & \textbf{Number of authors (~author type(s)~)} & \textbf{Number of articles} \\\hline\hline
True News  & one author ({\it True-news authors}) &   37 \\\hline
True News  & one author ({\it Fake+True news authors}) &   18 \\ \hline 
True News  & multiple authors ({\it True-news authors}) &   59 \\\hline
True News  & multiple authors  ({\it Fake+True news authors}) &   6 \\\hline
Fake News  & one author ({\it Fake news authors}) &  23 \\\hline
Fake News  & one author ({\it Both news authors}) &   18 \\ \hline 
Fake News  & multiple authors ({\it Fake news authors}) &  5\\\hline
Fake News  & multiple authors ({\it Both news authors}) &   6 \\\hline
\end{tabular}
\label{tab:distAuthors}
\end{table}

Not all news articles in our data had information on their publication date. As articles with no publication date could not help with historical analysis, we filtered them, reducing our data to 289 news articles. 
Among these 289 articles, we 
focused on authors 
of at least two news articles, which resulted in 69 authors 
 of 163 news articles. 
 For each author, their published news articles were sorted chronologically and we analyzed whether they contradicted their past behavior anytime in the future. We only found 11 authors that contradicted, i.e., either they had fake news in the past and were associated with true news in future, or they posted true news in the past but were associated with fake news in the future. However, the majority of authors (84\%) showed consistent behavior. 
 Thus, past information on authors' credibility provides insights on
 the credibility of other articles authored by them.

\section{Content Credibility}
\label{sec:content}
Next, we discuss credibility based on the content of the news. Our goal 
is to explore how various characteristics of a news article content (which includes
the text from
the title and
the body of 
an article) can help assess its credibility. These characteristics are compared 
for fake and true news articles. Previous research on credibility on Twitter~\cite{gupta2014tweetcred,castillo2011information,odonovan2012credibility,morris2012tweeting} has shown that there exist various indicators of credibility within content. Here, we search for such indications of credibility in the following: (1) sentiments expressed, (2) domain expertise in the article, (3) arguments used, (4) text readability, (4) characters, words, and sentences used, and (5) typos.

\subsection{Credibility Signals in Sentiments}
Castillo et al.~\cite{castillo2011information} identified connections between sentiments expressed and credibility, whereas O'Donovan, et al.~\cite{odonovan2012credibility}, found that positive sentiments may not indicate credibility in tweets. Such studies encouraged us to study the relationship between credibility and sentiments in news, answering questions such as: 
\begin{compactitem}
\item Are sentiments expressed in  fake news different from those in true news?
\item Is fake news written with negative, neutral, or positive sentiments? 
\item Can sentiments 
 help infer credibility?\vspace{1mm}
\end{compactitem}

For sentiment analysis, we compute sentiment intensity (a numeric value) for each sentence in the news articles using VADER~\cite{hutto2014vader} sentiment analyzer available in NLTK (natural language toolkit)~\cite{bird2009natural}. Using the standard threshold ~\cite{hutto2014vader}, three labels are assigned to each sentence: positive, negative, or neutral. We represent each news article with the three fractions of negative, positive, and neutral sentences in the article, e.g., $\frac{\mbox{\scriptsize number of positive sentences}}{\mbox{\scriptsize total  number of sentences}}$. Some statistics on such proportions of each type of sentiment in fake and true news are in Table \ref{tab:sentiment}.

\begin{table}[t]
\centering
\caption{Sentiment Proportions in News}
\begin{tabular}{|l|cc|cc|cc|}\hline

{\bf News} & \multicolumn{2}{c|}{\bf Positive} & \multicolumn{2}{c|}{\bf Neutral} & \multicolumn{2}{c|}{\bf Negative} \\
\textbf{Type} &     mean & median &    mean & median &     mean & median \\
\hline\hline
Fake News &     0.34 &   0.32 &    0.28 &   0.27 &     0.38 &   0.38 \\\hline
True News &     0.33 &   0.33 &    0.34 &   0.31 &     0.34 &   0.33 \\\hline
\end{tabular}
\label{tab:sentiment}
\end{table}

The mean and median values in Table \ref{tab:sentiment} show that (1) proportion of neutral sentiments is slightly higher in true news compared to fake news and (2) negative sentiments proportions are higher in fake news compared to true news. However, it is still difficult to infer whether sentiments are good indicators of credibility. For further analysis, we explored whether sequence of expressed sentiments in articles differ, i.e. is fake news more likely to have sequences of sentences with positive sentiments followed by other sentences with positive sentiments? There are 9 (3$\times$3) possible types of sequences that one can get with positive, negative, and neutral sentences. We label each sentence pair as one of these types. The mean and median proportion for each one of these nine types in both fake and true news are provided in Table \ref{tab:sentSeq}.

\begin{table}[t]
\centering
\caption{Statistics on sequences of sentences with different sentiments}
\begin{tabular}{|l|c|c|c|c|}\hline

\multirow{2}{0.68\textwidth}{\bf Sentence sequence} & \multicolumn{2}{c|}{\bf Mean} & \multicolumn{2}{c|}{\bf Median}\\
                                                                                        & \makecell{Fake\\News}& \makecell{True\\News} & \makecell{Fake\\News} & \makecell{True\\News}\\  
\hline \hline
Positive sentence followed by a positive sentence &  0.14 &  0.13 &   0.1 &  0.11 \\
Positive sentence followed by negative sentence &  0.10 &  0.09 &  0.10 &  0.09 \\
Positive sentence followed by neutral sentence &  0.09 &  0.11 &  0.08 &  0.10 \\
Negative sentence followed by positive sentence &  0.11 &  0.09 &  0.10 &  0.09 \\
Negative sentence followed by negative sentence &  0.16 &  0.14 &  0.13 &  0.11 \\
Negative sentence followed by neutral sentence &  0.11 &  0.12 &  0.10 &  0.11 \\
Neutral sentence followed by positive sentence &  0.08 &  0.09 &  0.07 &  0.10 \\
Neutral sentence followed by negative sentence &  0.09 &  0.08 &  0.09 &  0.07 \\
Neutral sentence followed by neutral sentence &  0.09 &  0.13 &  0.06 &  0.10 \\\hline

\end{tabular}
\label{tab:sentSeq}
\end{table}

Table \ref{tab:sentSeq} shows that positive sentences that are immediately followed by neutral sentences are more in true news compared to fake news. In fake news, negative sentences followed by other sentiments 
occur more often than in 
true news. Overall, the uniformity of values in Table \ref{tab:sentSeq} on sentiment sequences in news articles suggests that sentiment sequences may be weak indicators of credibility. In contrast to earlier findings ~\cite{castillo2011information,odonovan2012credibility}, 
our results show that relying on sentiments alone may provide only a weak indication of credibility. 

\subsection{Credibility Signals in Domain Expertise}
Research has shown that the presence of signal words and expert sources enhance credibility~\cite{wogalter1999effect}. As our data was collected during the U.S. 2016 presidential election, we looked into the use of words from NCSL (National Conference of State Legislatures), which included 150 words from \url{https://bit.ly/1iMTzXa}. We studied 
whether there exist
differences between fake news and true news in terms of 
usage frequencies of these words. We found that the average number of words from the NCSL word list were 4.37 in fake news and 7.46 in true news. The medians
for the number of words were 3 for fake news,
and 4 for true news. 
The Shapiro-Wilk test on the number of NCSL words on both fake and true news had a $p$-value of $<0.05$, showing the sample is not normally distributed on number of words for both types of news. With a small difference of one word, it is difficult to argue the importance of domain words/phrases, so, we further looked into distinct words that are present in one type of news and not in the other, shown in Table \ref{tab:NCSL}. 

\begin{table}[t]
\centering
\caption{List of NCSL words in fake and true news.}
\begin{tabular}{|p{0.25\textwidth}|p{0.72\textwidth}|}
\toprule
Words in fake news, but not in true news & \texttt{petition}, \texttt{legislator}, \texttt{impeachment}, \texttt{adhere} \\\hline
Words in true news, but not in fake news &  \texttt{fiscal}, \texttt{calendar}, \texttt{precedent}, \texttt{bipartisan}, \texttt{convene}, \texttt{interim}, \texttt{caucus}, \texttt{nonpartisan}, \texttt{statute}, \texttt{decorum}, \texttt{veto}, \texttt{repeal}, \texttt{constituent}, \texttt{chamber} \\
\bottomrule
\end{tabular}
\label{tab:NCSL}
\end{table}

Later in our experiments, we will show how the occurrence of words shown in Table \ref{tab:NCSL}, in addition to other information, allows one to detect fake news.

\subsection{Credibility Signals in Argumentation}
To build strong arguments in a news article, 
  one can rely on providing data and references. 
A greater occurrence of numbers or digits may 
 indicate that a news article is well-researched, containing verifiable data; similarly, the occurrence of hyperlinks and URLs may indicate citations 
 suggesting that an article is supported by external sources. 
 The connections between URLs and credibility have been studied earlier on tweets~\cite{castillo2011information,gupta2014tweetcred}. Similarly, the findings from Koetsenruijter~\cite{koetsenruijter2011using} suggests that the presence of numbers in an article
  conveys credibility. Hence, we studied whether there is a difference between fake news and true news in terms of numbers of URLs. 
We found that only 18 news articles 
contained URLs, of which 7 were fake news and 11 were true news. With presence of URL in small proportion of dataset, it is difficult to assess credibility strength based on this feature. Table \ref{tab:digits} shows the distribution of number of digits used in fake and true news. Our findings suggest that there are differences between fake news and true news based on the use of numbers in the news content, 
 and that it is likely that true news is supported with facts that include numbers. 

\begin{table}[t]
\centering
\caption{Distribution of digits in fake and true news}
\begin{tabular}{|c|c|c|c|}\hline
\multicolumn{2}{|c|}{\bf Mean} & \multicolumn{2}{c|}{\bf Median}\\
  Fake News &  True News &  Fake News &  True News\\ \hline\hline
 490.82 &	739.33	& 424 &	461 \\\hline
\end{tabular}
\label{tab:digits}
\end{table}

The Shapiro-Wilk test gave a $p$-value of $<0.05$, showing that the sample is not normally distributed on number of words for both type of news. The Mann-Whitney U test shows that there is a difference in medians with a $p$-value of $0.011$, i.e., the 
 greater occurrence of digits in news articles indicates 
credibility.

\subsection{Credibility Signals in Readability}
The study by Horne et al.~\cite{horne2017just} suggested readability as an important feature to distinguish fake news from true news. To compare readability differences between fake and true news, we used the Flesch-Kincaid reading-ease test on the text of the news. The mean readability scores were found to be 67.32 for fake news and 65.30 for true news. Similarly, the median scores are 68.33 and  65.38 for fake news and true news, respectively. Contrary to our expectation, fake news readability was higher than that of true news. This raises a series of other interesting questions such as: 

\begin{compactitem}
    \item Is fake news more readable?
    \item Is ease of reading why users engage more with fake news than true news?
\end{compactitem}

Further analysis of the news content may reveal insights on such questions. The Shapiro-Wilk test, with $p$-value $<0.05$ indicates that the sample is not normally distributed. The Mann-Whitney U test with a $p$-value of $0.02$ shows differences in medians and that poorer 
readability may indicate credibility.

\subsection{Credibility Signals in Characters, Words, and Sentences}
In TweetCred system~\cite{gupta2014tweetcred}, the number of characters and number of words were among the important features to evaluate credibility of tweets. Earlier work has also shown that tweet length is one of the indicators for credibility~\cite{gupta2014tweetcred,odonovan2012credibility}. Hence, the news content length may also be an indicator for the credibility. The mean and median of words in title and text of the news are shown in Table \ref{tab:numWords} along with the number of words per sentences.  
\begin{table}[t]
\centering
\caption{Distribution of words in title and text}
\begin{tabular}{|l|cc|cc|cc|cc|}\hline
{} & \multicolumn{2}{c|}{\bf \# words in title}& \multicolumn{2}{c|}{\bf \# words in text} & \multicolumn{2}{c|}{\bf \# sentences} & \multicolumn{2}{c|}{$\frac{\mbox{\bf \# words}}{\mbox{\bf \# sentences}}$} \\
\bf News Type &            mean & median &           mean & median &              mean & median &            mean & median \\
\hline\hline
Fake news &           14.16 &   14.0 &         490.82 &  424.0 &             19.44 &   16.0 &           26.80 &  25.32 \\\hline
True News &           12.09 &   11.0 &         739.33 &  461.0 &             26.59 &   17.0 &           27.69 &  27.50 \\ \hline
\end{tabular}
\label{tab:numWords}
\end{table}
Table \ref{tab:numWords} shows that fake news text is shorter in terms of number of words compared to true news. Similarly, the length of true news
 articles and 
their number of sentences were found to be higher on average compared to fake news.

In similar research, it has been shown that presence of special characters, e.g., colon~\cite{gupta2014tweetcred}, exclamation mark, and question mark~\cite{castillo2011information}, can help assess credibility. To check if fake news contains more special characters compared to true news, we select {\tt!}, {\tt\#}, {\tt\$}, {\tt\%}, {\tt*}, {\tt+}, {\tt-}, {\tt?}, {\tt@}, {\tt|}  as special characters and count their occurrences in each type of news. Table \ref{tab:special} provides the mean and median of number of characters and special characters in news text, indicating that special characters are more often observed in true news.
  
\begin{table}[t]
\caption{Distribution of characters and words}
\centering
\begin{tabular}{|l|cc|cc|cc|}\hline
{\bf } & \multicolumn{2}{c|}{\bf \# characters} & \multicolumn{2}{c|}{$\frac{\mbox{\bf \# characters}}{\mbox{\bf \# words}}$} & \multicolumn{2}{c|}{\bf \# special characters} \\
{\bf News Type} &           mean &  median &             mean & median &              mean & median \\
\hline\hline
Fake news &        2052.76 &  1803.0 &             4.23 &   4.20 &              6.79 &    4.0 \\\hline
True news &        3144.28 &  1899.0 &             4.27 &   4.27 &             11.04 &    6.0 \\ \hline
\end{tabular}
\label{tab:special}
\end{table}

\subsection{Credibility Signals in Typos}

As suggested by Morris et al.~\cite{morris2012tweeting}, the use of standard grammar and spellings enhances credibility. So, we checked if there are significant differences in typographical errors between fake and true news. To find if there are more typos in fake news compared to true news, we used the words from the NLTK corpus, containing 235,892 English words. For each news content, we counted the number of words with typos and normalized it by the total number of words in the content. Words from both content and NLTK corpus were lower-cased before checking for typos. Contrary to our expectation, we found on average 0.19 and 0.22 typos in fake news and true news, respectively. The median number of typos were 0.20 for fake news and 0.21 for true news. The Shapiro-Wilk test indicated that the sample is not normally distributed, and using Mann-Whitney U test, we obtained a $p$-value $<  0.001$, showing that typos may 
indicate credibility.

\section{Results}
\label{sec:result}
Based on our discussion on credibility aspects in Sections \ref{sec:source} and \ref{sec:content}, we built different fake news prediction models to predict fake news. From the attributes discussed, we obtained 26 features in the following categories:
 
\begin{itemize}
\item Number of authors in the news;
\item Sentiments (counts of positive, neutral, and negative sentiments in text, and sequence of sentiments);
\item Number of NCSL words that are only present in fake news;
\item Number of NCSL words that are only present in true news;
\item Flesch-Kincaid reading-ease score; 
\item Number of words in the title;
\item Number of characters, special characters, words, sentences, digits, and typos; 
\item Words per sentences; 
\item Characters per words; and 
\item Past history of the author.

\end{itemize}
Using the above features, we trained a fake news prediction model with seven different classifiers (to account for learning bias) using scikit-learn package~\cite{scikit-learn} and ten-fold cross validation. The $F_1$-scores for these classifiers are in Table \ref{tab:perf1}, where we found that the logistic regression and linear Support Vector Machines performed well among different classifiers. The best classification was achieved by logistic regression, with an 0.80 average $F_1$-macro score.\vspace{1mm}
\begin{table}[t]
\centering
\caption{Average $F_1$ scores for all features}
\begin{tabular}{|l|c|c|c|}\hline
\bf Classifier &  \bf $F_1$-micro  & \bf $F_1$-macro  &  \bf $F_1$-weighted  \\
\hline\hline
SVM (RBF Kernel)                          &          0.74 &          0.74 &             0.74 \\
Linear SVM                   &          0.79 &          0.79 &             0.79 \\
Logistic Regression                     &          {\bf{0.80}} &          {\bf{0.80}} &             {\bf{0.80}} \\
Random Forest                &          0.76 &          0.76 &             0.76 \\
AdaBoost                    &          0.74 &          0.74 &             0.74 \\
Naive Bayes        &          0.69 &          0.69 &             0.69 \\
Gradient Boosting Decision Tree &          0.77 &          0.77 &             0.77 \\ \hline
\end{tabular}
\label{tab:perf1}
\end{table}

\noindent \textbf{Comparing Source-Credibility and Content-Credibility.} We studied the importance of assessing each type of credibility (source and content) 
by predicting fake news independently using each category of features. For source-credibility, we only considered three features: number of authors, 
as well as the numbers of past fake and true news stories authored by them in the past. Surprisingly, with these three features, we find that the classification performance does not degrade much, as shown in Table \ref{tab:perf2}. However, the best classifier was then 
AdaBoost, which indicates that the 
classifier performance is feature-dependent. Similarly, when using the 23 content-credibility features, the best $F_1$-score achieved was 0.68, less than when using only source-credibility features, which achieved 0.77. The results are shown in Table \ref{tab:content}.

\begin{table}[t]
\centering
\caption{Average $F_1$ scores obtained by source-credibility features}
\begin{tabular}{|l|c|c|c|}\hline
\bf Classifier &  \bf $F_1$-micro  & \bf $F_1$-macro  &  \bf $F_1$-weighted  \\
\hline\hline
SVM (RBF Kernel)                        &          0.75 &          0.75 &             0.75 \\
Linear SVM                   &          0.75 &          0.75 &             0.75 \\
Logistic Regression                   &          0.75 &          0.74 &             0.74 \\
Random Forest                &          0.76 &          0.76 &             0.76 \\
AdaBoost                    &          \textbf{0.77} &          \textbf{0.77} &             \textbf{0.77} \\
Naive Bayes        &          0.75 &          0.75 &             0.75 \\
Gradient Boosting Decision Tree &          \textbf{0.77} &          0.76 &             0.76 \\ \hline
\end{tabular}
\label{tab:perf2}
\end{table}

\begin{table}[t]
\centering
\caption{Average $F_1$ scores obtained by content-credibility features}
\begin{tabular}{|l|c|c|c|}\hline
\bf Classifier &  \bf $F_1$-micro  & \bf $F_1$-macro  &  \bf $F_1$-weighted  \\
\hline\hline
SVM (RBF Kernel)                             &          0.64 &          0.63 &             0.63 \\
Linear SVM                   &          \textbf{0.68} &          \textbf{0.68} &             \textbf{0.68} \\
Logistic Regression                   &          0.67 &          0.67 &             0.67 \\
Random Forest                &          0.63 &          0.63 &             0.63 \\
AdaBoost                    &          0.60 &          0.60 &             0.60 \\
Naive Bayes        &          0.58 &          0.57 &             0.57 \\
Gradient Boosting Decision Tree &          0.65 &          0.65 &             0.65 \\ \hline
\end{tabular}
\label{tab:content}
\end{table}

By comparing the performance of source-credibility and content-credibility features, we find that assessing source credibility plays a stronger role in detecting fake news. Adding content-credibility features with source-credibility features can further improve fake news detection.\vspace{1mm}

\noindent \textbf{Feature Importance Analysis.} We also identified the most important features that can capture credibility in news. 
While there can be various combinations of features to search for the optimal features, we combined both the feature selection and a hand-tailored approach (testing with trial and error), which led to 13 features with the best $F_1$ score:
number of authors in the news and past history of authors, presence of domain words, readability,  number of words, characters, special characters, and typographical errors. Table \ref{tab:features} shows that all classifiers performed best with these selected features, even better than using the original 26 features. Also, 
features that were found to be of least importance were sentiments and count of digits in the text.

Comparing Table \ref{tab:perf1} and Table \ref{tab:features}, we can observe that using these 13 selected features, all the classifiers perform better than using the original 26 features. While our results did not outperform all other models as discussed in \cite{zhou2019fake}, where best model had a 0.892 $F_1$ score, our model used comparatively 
fewer and new features compared to the models discussed in the work.

\begin{table}[t]
\centering
\caption{Average $F_1$ score with only 13 features}
\begin{tabular}{|l|c|c|c|}\hline
\bf Classifier &  \bf $F_1$-micro  & \bf $F_1$-macro  &  \bf $F_1$-weighted  \\
\hline\hline
SVM (RBF Kernel)                       &          0.77 &          0.77 &             0.77 \\
Linear SVM                   &          {\bf{0.80}} &          {\bf{0.80}} &             {\bf{0.80}} \\
Logistic Regression                    &          {\bf{0.80}} &          {\bf{0.80}} &             {\bf{0.80}} \\
Random Forest                &          0.77 &          0.77 &             0.77 \\
AdaBoost                    &          0.77 &          0.77 &             0.77 \\
Naive Bayes        &          0.75 &          0.75 &             0.75 \\
Gradient Boosting Decision Tree &          0.77 &          0.77 &             0.77 \\ \hline
\end{tabular}
\label{tab:features}
\end{table}

\begin{table}[t]
\centering
\caption{Average $F_1$ score using 13 features on separated datasets}
\begin{tabular}{|l|c|c|}\hline
\bf Classifier   & \bf  \makecell{$F_1$-macro\\(Politifact)} & \bf \makecell{$F_1$-macro\\(Buzzfeed)} \\
\hline\hline
SVM (RBF Kernel)                            &          0.79 &            0.75 \\
Linear SVM                    &          \textbf{0.82} &             \textbf{0.77} \\
Logistic Regression                  &          \textbf{0.82 }&             0.76  \\
Random Forest                &          0.77 &             0.72 \\
AdaBoost                    &          0.78 &            0.67\\
Naive Bayes        &          0.79 &             0.69 \\
Gradient Boosting Decision Tree &          0.80 &              0.74 \\ \hline
\end{tabular}
\label{tab:separate}
\end{table}

With only 3 source-credibility features on Politifact data, the best classifier achieved an average $F_1$-macro score of 0.83 and with only content-credibility features, best score was 0.66 (see Table \ref{tab:separate}). This observation shows that content-credibility has very 
little to add to the performance of fake news prediction in the data. Similarly, for Buzzfeed news data, the best classifier was able to obtain an average $F_1$-macro score of 0.76 with only source-credibility features,  whereas with content-credibility features it obtained 0.66. 
Thus, adding content-credibility features only slightly improved the performance. Our content-credibility features are comparatively 
fewer than earlier studies, so we 
emphasize 
our findings with source-credibility features, which we did not find in earlier research.

\section{Conclusion}
\label{sec:conclusion}

We have analyzed credibility of news, 
emphasizing features related to source and content of the articles. Our results based on source of the news (Section \ref{sec:source}) show that number of authors of the news is a strong indicator of credibility.  We found that when the news article has no authors, it is more likely to be fake news. Our findings on collaboration of authors suggests that authors who are engaged in true credible news are less likely to collaborate with authors who are associated with fake news. This indicates that for a news article with multiple authors, by knowing the credibility of one 
author, we can infer the credibility of the news as well as other coauthors. Furthermore, we found that authors' affiliations with well-recognized organizations can be a signal for credibility. The results also suggest that credibility history of authors can provide insights on credibility of other articles from the same author.

Similarly, we investigated credibility based on various content-related aspects of the news (Section \ref{sec:content}). The results show that sentiments expressed in news articles are weak indicators of credibility. We observed that the use of numbers in true news articles
 occurred more often than in 
 fake news, 
 perhaps because true news is supported with facts that include numbers. Comparing the number of words and sentences in true news and fake news showed that on average, true news had more words and sentences than fake news. Surprisingly, we observed more typos in true news than in fake news. Our analyses also showed that domain expertise on topics discussed in news can enhance fake news detection.

After analysis of individual features, we used our findings to build predictive models to detect fake news. The $F_1$-score of 0.80 obtained by predictive models built with source-credibility features show that with a small number of features, one can still can detect fake news reasonably well. 
Using fewer features can lead to less complex models. Hence, our simple approach provides a straightforward fake news detection framework with a few features that can quickly detect fake news.

Stronger conclusions require further research on additional machine learning features, other predictive models, and datasets. We have not yet explored word-based sentiments in our analysis, where one can consider negated positive words, or number of negative and positive words in sentences. Another avenue to explore is to study sentiments in paragraphs, which may show less variation compared to our results. Furthermore, the news content can include images (and other media), as well as the number of user interactions, which may provide more insights on the differences between fake and true news. 

\bibliographystyle{splncs04}
\bibliography{reference}

\end{document}